# INCORPORATING ENSEMBLE AND TRANSFER LEARNING FOR AN END-TO-END AUTO-COLORIZED IMAGE DETECTION MODEL

[1]**AHMED SAMIR RAGAB**, [1]**DR. SHEREEN ALY TAIE**, [1]**DR. HOWIDA YOUSSRY ABDELNABY**

E-mail: as135@fayoum.edu.eg, sat00@fayoum.edu.eg, hya11@fayoum.edu.eg

[1]Department of Computer Science, Faculty of Computers and Artificial Intelligence, Fayoum University, Fayoum, Egypt

## ABSTRACT

Image colorization is the process of colorizing grayscale images or recoloring an already-color image. This image manipulation can be used for grayscale satellite, medical and historical images making them more expressive. With the help of the increasing computation power of deep learning techniques, the colorization algorithm's results are becoming more realistic in such a way that human eyes cannot differentiate between natural and colorized images. However, this poses a potential security concern, as forged or illegally manipulated images can be used illegally. There is a growing need for effective detection methods to distinguish between natural color and computer-colorized images. This paper presents a novel approach that combines the advantages of transfer and ensemble learning approaches to help reduce training time and resource requirements while proposing a model to classify natural color and computer-colorized images. The proposed model uses pre-trained branches VGG16 and Resnet50, along with Mobile Net v2 or Efficientnet feature vectors. The proposed model showed promising results, with accuracy ranging from 94.55% to 99.13% and very low Half Total Error Rate values. The proposed model outperformed existing state-of-the-art models regarding classification performance and generalization capabilities.

## 1. INTRODUCTION

Image colorization is the process that adds color to a grayscale image to obtain a realistic color image. Doing this manually consumes effort and time; thanks to machine learning techniques, the colorization process could be as simple as a button click.

There are three main colorization methods, Scribble-based, Example-based (reference-based), and fully automatic approaches.

Scribble-based methods [1]-[5] is a supervised technique in which the user begins assigning colors to pixels in the grayscale image and then assumes that the neighboring pixels with similar intensities should have similar colors; the result is a color image. This method is usually accompanied by trial and error to obtain satisfactory results; this relies on, is limited by the user's experience, requires a large number of experiments to achieve good performance, and thus is instead a time-consuming process.

Example-based (reference-based) algorithms [6]-[8] are also supervised techniques that require the user to supervise the system by providing reference color images semantically similar to the greyscale image. The system then transfers the colors in the reference color image to the target greyscale image by searching for similar patterns/objects.

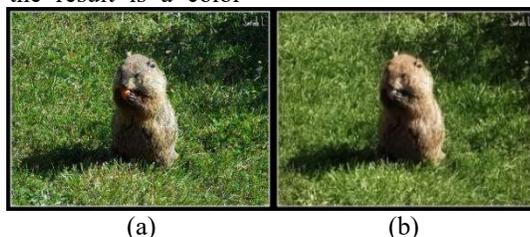

(a)              (b)

*Figure 1: (a) is the Computer Colorized Image CCI generated by the colorization method proposed in [9] from the grayscale version of (b), (b) is a Natural Color Image NCI Picked from ImageNet [24]*

The performance then depends on the quality of the reference image, and selecting a suitable reference image may be difficult.

In contrast to the above methods, the Fully automatic colorization methods [9]-[12] are unsupervised techniques that do not need a user interaction or an example image. Training a neural network to predict per-pixel color histograms utilizing low-level and semantic representations. Using a grayscale image as input to a trained, fully automatic colorization neural network outputs a color image without interaction. These method results are plausible enough to be detected as Computer Colorized Images (CCI) by the human eye. The last method (automatic colorization) gets the most research interest, and the proposed models' purpose is to detect the resultant colorized images using automatic colorization approaches.

The computer-colorized images can be used for entertainment and other non-high-importance fields without affecting legal or security decisions. This includes colorizing old grayscale photos and recoloring personal images to vibrant colors other than the original colors.

Legal or security usages of color images will require a technique to detect whether it is a computer-colorized or a raw image.

Using a computer-colorized image as a raw image in the healthcare, criminal identification, and urban planning fields could cause a wrong decision. Color image usage in the legal and security fields is becoming tremendous, making the colorized image detection step in the process of color image usage in such fields a vital step in making the right decision.

Image Colorization can be categorized as a passive forgery pixel-based image tampering technique that can be detected using statistical analysis and semantics of the image properties and features. Other tampering techniques, such as camera-based, physical-based … etc., can be detected using different methods.

**The contributions of this paper are:**
- An end-to-end framework to classify computer-colorized and natural color images with high accuracy and best-known generalization performance.
- A new training and testing approach uses 1:3, one natural color-to-three computer-colorized images for training and testing the proposed model.

The performance of the proposed model is tested and compared with the state-of-the-art methods.

The rest of the paper is organized as follows. Section 2 presents the related work. Section 3 introduces the proposed model. Section 4 shows the Experiment and Results. Section 5 Discussions to compare the results to the other models' performance. Finally, Section 6, Conclusion, summarizes the paper and discusses future work.

## 2. RELATED WORK

In 2018, Y.Guo et al. [13] were the first to publish a detection method for colorized images, followed by other researchers who tried to develop Y.Guo et al.'s methods (histogram, feature encoding) using other algorithms instead of support vector machine.

Then newly published algorithms modified some hyper-parameters or model structures to enhance the detection performance and accomplish better results than Y.guo's, with some drawbacks like the increased training time or decrease in the model's generalization performance.

While all related papers used Ctest10K [10] dataset and the same accuracy metrics HTER, results are combined into one table at the end of this section. This gives a summarized overview of all related work.

The detection approaches will be categorized into three categories; 1) Hand-crafted features approaches, 2) Learned Feature approaches, and 3) Merging both approaches (handcrafted and learned features).

**Hand-Crafted Features Approaches**

Y.Guo et al. [13] were the first to propose a fake colorized image detection technique; they used two approaches to detect colorized images (FCID-HIST Histogram based, FCID-FE feature encoding based). They used a support vector machine (SVM) classifier to calculate the statistical difference between natural images and auto-colorized mage properties (Hue - Saturation, Dark - Bright channels). As they use handcrafted methods, the detection is built on the prior knowledge observed from data. This leads to a performance drop when training and testing images produced by different colorization methods.

The FCID-FE method performs better than the FCID-HIST method; FCID-FE can be modified using Fisher vectors and other encoding methods other than GMM. With a more in-depth study of the common characteristics of the SOTA automatic colorization methods, FCID-FE will perform better in generalizing the solution.

Saurabh Agarwal et al. [20] used the local binary pattern (LBP) operator that is often used in feature extraction and the Linear Discriminant Analysis (LDA) classifier for binary classifications; both are used together to detect CCIs. Saurabh Agarwal et al. found that their LBP operator method performs better with lower computational cost than [13], using mainly the dataset ImageNet test10K of [9].

Yangxin Yu. et al. [23] used the LCA feature shown in the natural images taken by digital cameras because of the imperfection of the physical property of the camera lens to differentiate the Nis from CCIs, which have been colorized using computer algorithms. Their proposed method performs better than [13] and moderates better than [16], who used deep learning low convolutional layers. While [17] used a deep learning model with high convolutional layers that performed better than using the LCA feature to detect CCIs.

Zhang et al. [40] found that the colorization techniques leave some degrees of damage to the texture of the original image as an image reconstruction problem during the encoding and decoding of the colorized image features. They proposed a SVM to classify colorized and natural images using the LBP operator. They analyzed the LBP operator of the RGB, HSV, and YCbCr spaces to find the most discriminative texture artifacts in which color space to be used to classify the colorized images. Their analysis shows that the chrominance space's texture information discriminates most to the colorized images. They used the ImageNet Validation dataset[24] and COCO validation dataset [41] for their experiment; their proposed method performance achieves better accuracy than [23] and [32] as they used the chrominance features that are more representative of the common features of colorized images from different sources.

### Learned Features Approaches

Long Zhuo et al. [14] used steganalysis algorithms to detect colorized images. They assumed that when fully automatic colorization methods reconstruct the red, green, and blue channels from a single grayscale template, it is inevitable that artifacts are introduced in the inherent statistical properties among RGB channels of the colorized image. The task of true-color image steganalysis is to expose the artifacts hidden among the RGB channels of the stego images. The task of detecting fake colorized images is similar to that of true-color image steganalysis, so they apply a true-color image steganalyzer to detect fake colorized images generated by fully automatic colorization methods. Long Zhuo used a Tensorflow network named WISERNet (Wider Separate-then-reunion Network), a deep learning-based data-driven color image steganalyzer [15], and the dataset used by Y.Guo [13], the performance was better than of the handcrafted features methods FCID-FE and FCID-HIST proposed by Y.Guo.

Weize Quan et al. [16] tried to enhance Y.Guo et al. [13] results; Weize Quan used an end-to-end framework based on Convolutional Neural Network built on a modified BaseNet [28] architecture to learn informative and generic characteristics automatically between Natural Color Images (NCIs), and Computer Colorized Images CCIs that have been created by [9] - [11].

To improve their proposed work performance added a generalization capability to Y.Guo's work. Wieze Quan's model then outperforms Y.Guo's results. Weize Quan used a proposed Generalization approach by inserting negative samples that were automatically constructed from the available training samples to help enhance the network-training phase. They added a new branch to the network architecture borrowed from ensemble learning to combine multiple predictions of a set of individually trained classifier to extract more features.

Weize Quan et al. [17] then improved the generalization capability of the detection by editing their training phase and modifying WISERNet; they constructed negative samples through linear interpolation of paired natural and colorized images. Then progressively inserted these negative samples into the original training dataset and continued training; this enhanced training technique significantly improved the generalization performance of different CNN models used in the detection process but with a slight decrease in the classification accuracy.

Weize Quan et al. [19] found that the data preparation phase affects the improving generalization performance of the detection process, and the CCIs JPEG compression badly affects the performance of the generalization detection process in both Y.gou et al. [13] and Long Zhuo et al. [14]. Using WISERNet's first layer, they improve classification accuracy using CCIs of the same colorization algorithm in training and testing and generalization performance using CCIs of different colorization algorithms in testing and training. The CCIs they used are from the three-colorization algorithms [9], [10], [11], and their corresponding NCIs from ImageNet test10K of [10]. They found that to improve detection accuracy and generalization performance should opt out of all testing and training images with JPEG compression from the dataset used.

Ulloa C. et al. [21] proposed a custom NN model to detect CCIs. This model uses Images resulting from automatic and manual colorization algorithms, they compared their model with VGG16 [22], and VGG16 outperforms their model in terms of performance results with training time three times longer than their proposed custom model to be trained, Twelve times longer in inference time, their custom model is a better solution for high-volume image classification. They also found that using the transfer-learning-based model VGG16 outperforms all previously proposed models that use WISERNet [15].

**Merging handcrafted and learned features**

Yuze Li et al. [18] analyzed the statistical differences between CCIs and their corresponding NCIs to detect the significant differences; they also used the cosine similarity to measure the overall similarity of normalized histogram distribution of various channels for natural and CCIs to extract features for detection. They studied the statistical differences in the color distribution between NCI and CCI. They found that the RGB color space has redundant information that leads to insufficient feature differentiation, so they used RGB (red, green, blue) with HSV (hue, saturation, value) representations to extract more features.

Yuze li et al. used a modified DenseNet [27] called ColorDet-NN with the same dataset ctest10K [10], commonly used by the other approaches. The feature analysis they performed on the output images from the automatic colorization methods to determine the handcrafted features that commonly enable detecting CCIs, the most significant feature they found to have an effect in the detection process is the color saturation channels distribution between NCI and CCI. Their model results have better performance than Y.Guo's [13]. However, the performance is less when the training and testing image sets are drawn from different datasets, which leads to a drawback in terms of the generalization performance of their algorithm.

Bonthala Swathi et al. [37] proposed a deep neural network (CNN) with inputs HSV color image, the derived difference image inter-channel correlation, and the computed RGB version of the same image both as one input. The third input to the model is the illumination mapping of the RGB input image. The inter-channel correlation is the relationship between image channels (H, S, V) of the input image. Because these channels are not independent, recoloring an image could have an artifact detected as a feature to identify the colorized image. Illumination mapping is used to maintain the consistency of illuminant colors in an image; this consistency cannot be maintained for a colorized image. Bonthala Swathi et al. used inter-channel correlation and illumination mapping for detecting recolored images, as the correlation may be disrupted or altered after a recoloring process; besides the illumination mapping inconsistent of a recolored image, both with the original input image fused to train a convolutional neural network to classify colorized and normal images. They used MATLAB CNN with 16 layers with the VOC PASCAL2012 [38], and their model's accuracy was 100% without showing the size of the training, validation, and testing subsets. Also, the generalization performance of their model using different colorization algorithms and showing the model performance on each is not discussed. Bonthala Swathi, et al. experiment's results did not state the number of images used in training and testing their proposed method to be compared to the others.

Phutke et al. [39] observed from their proposed channel difference map of fake-to-fake and real-to-real images that the fake colorized images have blurred edges and fewer color shades. They proposed an auto-encoder based on the difference image regeneration followed by a fake colorized image detection framework; their architecture first concatenates the channel differences and then fed them to the classification framework (Dense module) to correlate the color and edge information from each channel for effective image regeneration, then a decoder module used to regenerate the input image back. After training this regenerate network, they used the transfer learning approach to use the trained regenerate encoder weights as initial weights for the proposed classification network denoted as DCDNet.

Shashikala s. et al. [42] used three corresponding 2D scaleograms for each HSV channel of the images, then three separate modified Densenet [27] classifiers trained to classify 2D scaleogram image NCI or CCI. The results of the preceding three layers are combined by an ensemble learning approach to calculate the probability of the image being NCI or CCI. They used 5000 NCI from ImageNet [24] and 5000 CCI from CTest10k[10].

The previously proposed methods that used the same metrics and dataset are compared in the following table (Table 1), each with its contribution, dataset, and detection algorithm used.

In Table (1), each paper by its references index with the corresponding dataset, detection model/algorithm, and a summary note about its accuracy is grouped to summarize all the previous work. This gives quick guidance when reading any of these papers from their source.

Table (1) lists All previous work that mostly used the dataset ImageNet ctest10K, with some modifications like JPEG Compression as [19] did. Others [18], [20], [21], [23] added datasets (Oxford buildings [25], CG-1050 [26], etc..) to be detected by their approaches or removed a subset of images from ctest10K as [16] did in the training phase.

*Table 1, Shows a detailed comparison of related work detection approaches with the used datasets, the model proposed, and a note on the resultant accuracy and performance.*

| Paper | Detection Model/Algorithm | Dataset | Research Findings |
|---|---|---|---|
| [13] | SVM | ImageNet ctest10K[10] | FCID-FE is better than FCID-Hist. |
| [14] | WiserNet [15] | ImageNet ctest10K Oxford building DS[25] | Accuracy better than [13] |
| [16] | DECNet: Modified BaseNet [28] | ImageNet ctest10K (removing gray and CMYK (900 images) CCIs from [10] &[11] Oxford building DS | Accuracy better than [13] |
| [17] | AutoNet: WiserNet[15] with trained weights | ImageNet ctest10K[10] | Negative sample insertion improved detection accuracy |
| [18] | ColorDet: DenseNet [27] with 4fouhidden layers | ImageNet ctest10K CCIs from [10] &[11] | Accuracy >88% (train & test using same colorization Technique) Accuracy >73% (train & test using different colorization Technique) Overall Acc. Better than [13] |
| [19] | Modified WiserNet[15] 30 - 5X5 SRM residual filters & 1[st] layer trainable | ImageNet ctest10K[10] -JPEG compression effect | JPEG compression affects the detection in [13] and [14] |
| [20] | Linear Discriminant Analysis (LDA) classifier | ImageNet ctest10K[10] & 1519 images from [12] & 5150 images from [29] & 1338 images from [30] & 10000 images from [31] | Better performance than [13] Using LBP (local binary pattern) operator. Low computational cost |
| [21] | Custom Model: 3Conv. Layers Transfer Learning Model: VGG-16 (13 Conv. Layers) | CG-1050 [26] Ctest10K[10] | Accuracy VGG16[22] has higher training(3x) and inference(12x) time |
| [23] | lateral chromatic aberration (LCA) Feature | ImageNet ctest10K Oxford building DS | Higher performance than [13] Moderate better than low CNN of [16], Fall behind the AutoNet proposed by [17] |
| [37] | CNN 16 Layers | VOC PASCAL2012 [38] | The size of the training, validation, and testing subsets not determined. Need to assess the generalization performance. |
| [39] | Channel Difference Map-based Auto-Encoder, DCDNet | ImageNet ctest10K[10] Oxford building DS[25] | Accuracy and generalization performance better than [13] & [14] |
| [40] | SVM | COCO validation dataset [41] | Accuracy better than [23] |
| [42] | Modified Densenet [27] | ImageNet ctest10K[10] | Accuracy better than [21] by 2.2% |

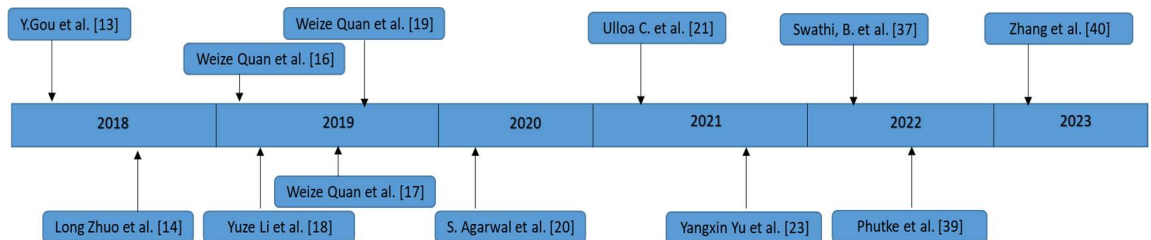

*Diagram 1 shows the timeline of the papers in Table 1*

The above diagram shows the related work timeline according to the publishing date.

Categorizing the proposed models of the related work papers are discussed into three categories, 1) Handcrafted features category uses the CCIs, and NCIs to extract statistically the features that can be used manually to differentiate CCIs from NCIs by providing them to the classification model. 2) Learned features approaches used automatically learned features extracted by Neural Network models named in front of each paper in table (1) with its architecture or the base network it has been built on with some layer modifications. 3) Mixed handcrafted and automatic learned features approaches used the statistical analysis of both CCIs and NCIs to get the most features found to differentiate, then feeding those features to their proposed model ColorDet that has been built on DenseNet [27] with four hidden layers, which provides a good technique to get the most benefits from the two previous categories. Despite their model performance's acceptable accuracy, the generalization performance degrades when the colorized images are from another colorization algorithm that their model trained on.

The notes on the accuracy comparison between the related work papers in Table (1) are based on the results stated in each.

The accuracy of the [13]'s method FCID-FE was first found to be better than FCID-Hist, then when others as [14], used a neural network instead of a support vector machine and with some modification in the training phase as a step of generalizing the detection accuracy found to be better than FCID-FE method of [13]. Weize Quan et al. [19] model's accuracy and generalization capability decrease when using CCIs with JPEG compression in training and testing datasets.

Using the related work papers that used the same dataset and listed their results in the next table to compare their performance in terms of HTER.

Table (2) shows the discussed related work proposed models as follows; BaseNet [28] is 8 Conv. Layers and a fully connected classifier (9 layers) that is inspired by DenseNet [27]. DecNet is a BaseNet With a newly inserted branch ([16] contribution). DecNet-i is the same as DecNet with enhanced training using the negative sample insertion technique. AutoNet is a WiserNet [15] with the first layer untrainable. AutoNet-i and WiserNet-i are for the enhanced training method by negative sample insertion using the model followed by "-i".

In Table (2), the Ma column denotes the fake colorized images resulting from the colorization method proposed by [9], Mb for fake colorized images from method [10]; Mc denotes fake colorized images from [11], training and testing with these three datasets interchangeably to assess the generalization performance. The bold results denote the best model accuracy when trained and tested using the corresponding datasets.

*Table 2, Compares SOTA models used to detect fake colorized images. Results are in half-total error rate (HTER), where lower is better.*

| Algorithm | Training | Ma [9] | | | Mb [10] | | | Mc[11] | | |
|---|---|---|---|---|---|---|---|---|---|---|
| | Testing | Ma [9] | Mb [10] | Mc [11] | Ma [9] | Mb [10] | Mc[11] | Ma [9] | Mb [10] | Mc [11] |
| FCID-HIST [13] | | 22.50 | 28.00 | 33.95 | 26.95 | 24.45 | 41.85 | 38.15 | 43.55 | 22.35 |
| FCID-FE [13] | | 22.30 | 23.65 | 31.70 | 25.10 | 22.85 | 34.25 | 38.50 | 36.15 | 17.30 |
| BaseNet [28] | | 0.56 | 10.57 | 10.62 | 31.65 | 0.19 | 6.16 | 13.93 | 1.91 | 0.72 |
| DecNet [16] | | 0.55 | 7.62 | 7.09 | 26.12 | **0.16** | 3.53 | 13.09 | 2.12 | 0.55 |
| DecNet-i [16] | | 1.03 | 5.09 | 4.13 | **4.41** | 0.85 | **1.60** | 2.83 | 1.77 | 0.98 |
| AutoNet [17] | | 0.56 | 10.57 | 10.62 | 31.65 | 0.19 | 6.16 | 13.93 | 1.91 | 0.72 |
| AutoNet-i [17] | | 1.02 | 6.94 | 5.12 | 5.13 | 0.94 | 1.92 | 3.33 | 1.75 | 1.14 |
| WISERNet [14] | | **0.29** | 2.21 | 10.74 | 33.30 | **0.16** | 7.88 | 5.80 | **0.59** | **0.36** |
| WISERNet-i [17] | | 0.98 | **1.22** | **2.29** | 4.74 | 0.94 | 2.04 | **2.46** | 1.08 | 0.98 |
| ColorDet-NN [18] | | 13.85 | 30.45 | 27.00 | 25.80 | 12.35 | 20.55 | 25.45 | 20.95 | 13.85 |

From the results, the deep learning models' detection accuracy outperforms all other models, considering that detection performance is the priority; in such cases, computational cost and training time are traded off for performance. Using transfer-learning VGG16 as the proposed model, their two model results are listed in the next table.

In Table (3), Ulloa, C et al. [21] compared their transfer learning VGG-16-based model with other state-of-the-art models using the difference between internal and external validation HTER (Half Total Error Rate); the [21] results show higher classification performance with the best generalization accuracy for VGG-16-based model.

*Table 3 shows that [21] VGG-16 achieves the best detection accuracy with the lower HTER Difference.*

| Algorithm | Dataset | Internal Validation | External Validation | HTER's Difference (External–Internal) |
|---|---|---|---|---|
| WISERNet [14] | [9]+[10]+[11] | 0.95 | 22.5 | +21.55 |
| WISERNet-i [17] | [9]+[10]+[11] | 0.89 | 4.7 | +3.81 |
| Custom model [21] | [9] + [26] | 9.00 | 16.0 | +7 |
| VGG-16-based model[21] | [9] + [26] | 2.60 | 2.9 | +0.3 |

The results above will guide the best approach to building the proposed models for the best classification and generalization accuracy.

This study will cover the related work gaps through related work evaluation by answering the research questions and objectives. This study will show the potential of ensemble learning and transfer learning approaches in detecting auto-colorized images with higher precision and generalization capabilities than all previously published methods.

### 3. PROPOSED MODEL:

The intuition of the proposed models is based on the use of transfer and ensemble learning together; using their benefits gives the models the advantages of high accuracy and less time and resources used for training.

The proposed models, as illustrated in fig. (1), fig. (2), and fig. (3), have two branches; the first is the pre-trained branch, either the VGG16 feature vector or the Resnet50 feature vector, both used to extract the basic features of the training datasets, the second branch is the Mobile Net v2 or Effecientnet feature vector, both trained to extract features from colorized and natural color images. Both branch's features are then concatenated to be the input for the Dense top layer of the models with its two neurons that are trained to perform classification of the two image classes based on the extracted features from the preceding two branches to get a final result of if the image is a Colorized or natural color image.

The proposed models' architecture comprises three phases, preprocessing, feature extraction, and classification phases, as illustrated in fig. (1), fig. (2), and fig. (3); each phase is discussed in detail in the following subsections.

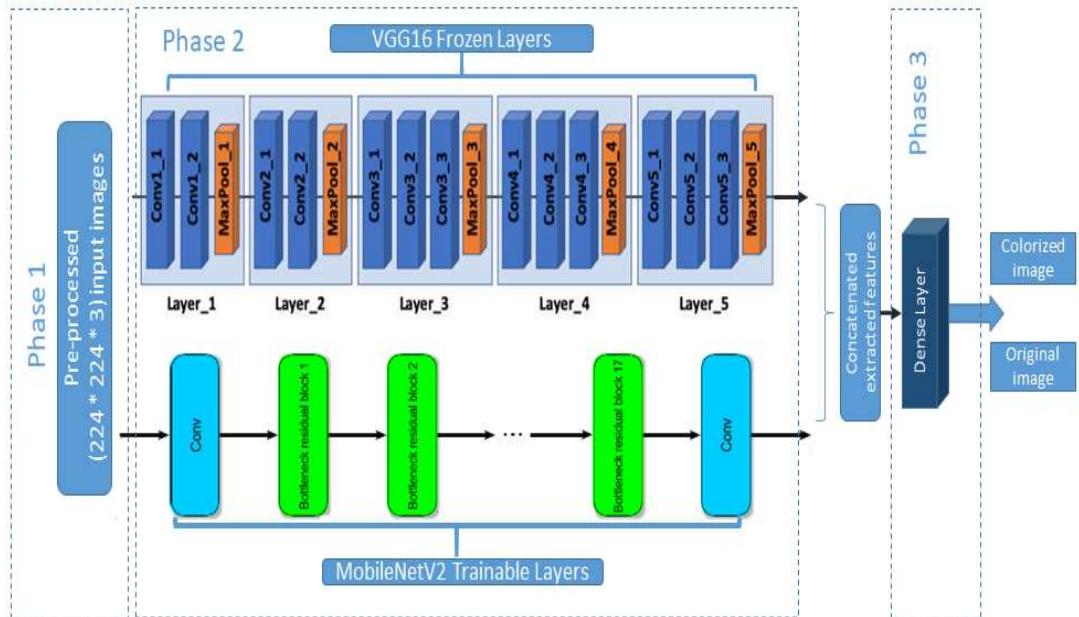

Figure 1. VGG16-based Model (1) architecture

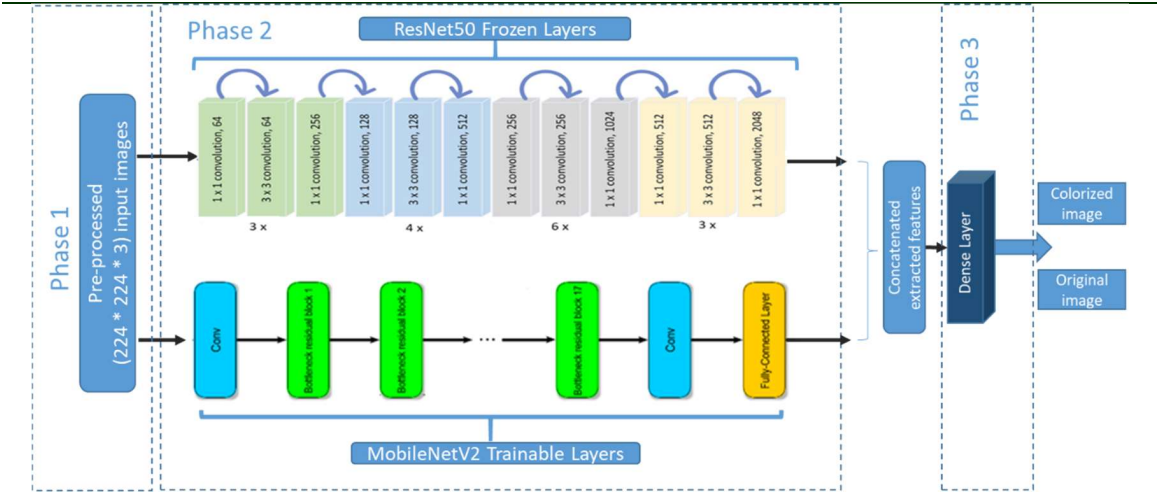

Figure 2. Resnet50-based Model (2) architecture

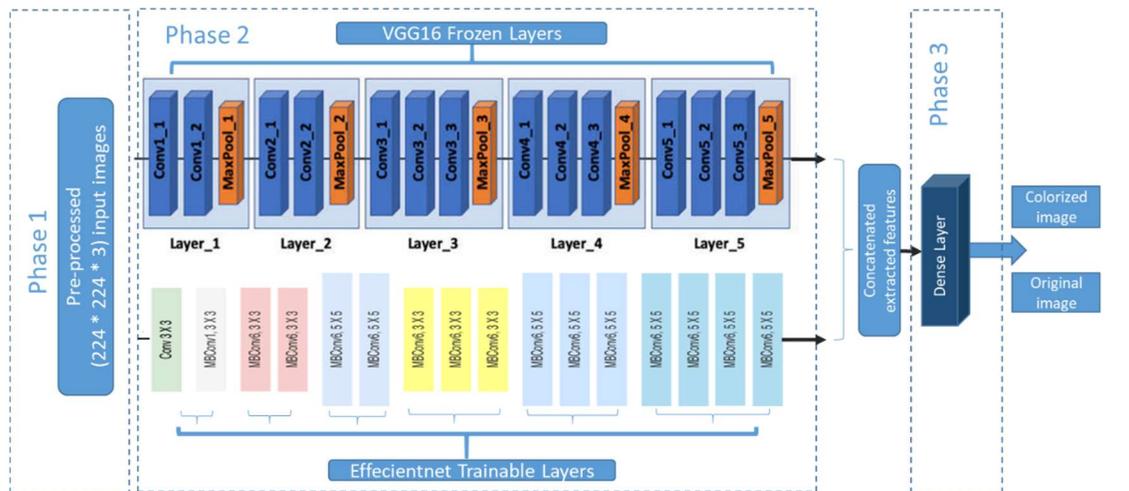

Figure 3. Vgg16-Effecientnet based Model (3) architecture

### 3.1. Preprocessing Phase

In this phase, the datasets used for training and testing are preprocessed the same way by resizing the images, normalizing pixel values, and shuffling. The preprocessing steps are as follows:

**Step1: Image resizing:**

To ensure uniform processing and optimize resource utilization during training and testing, the original sizes of the 10,000 natural images and their corresponding 30,000 colorized images of the three datasets DS1, DS2, and DS3 were resized to 224 x 224 pixels.

**Step2: Image normalization:**

To improve the model's robustness to variations in lighting, color, and other features, employing Min-Max normalization to scale the pixel values of all images in the dataset to a range of 0-1. This was achieved by implementing a normalization layer with a scaling factor of 1.0/255, which was applied to the entire dataset to ensure consistency in the normalization process.

**Step3: Image shuffling:**

To prevent overfitting and ensure that the model is exposed to diverse examples of each class, the images in the dataset were shuffled to distribute the classes randomly. This enhances the model's ability to learn from various examples and improves its generalizability.

### 3.2. Feature Extraction Phase

In the feature extraction phase, deep neural networks are employed to extract features relevant to the images' output labels. This approach was chosen over the use of handcrafted features.

Given the large number of images used for training and testing, and as part of the ImageNet dataset, which comprises 1 million images, utilizing transfer learning and fine-tuning techniques to use the features and weights obtained from a pre-trained model that had been trained on the ImageNet dataset.

The feature vector of a model is the base component used in ensemble learning architectures.

In a two-level deep learning model, the feature vector is referred to as the level-0 layer of the model. The feature vector is obtained by removing the top layer of the model; it is then used to extract the fundamental characteristics or features of the dataset images. These extracted features are then fed to the subsequent level, a custom level-1 layer that fits the number of classes to be classified; this custom level-1 layer will be trained to classify the two classes (colorized or original). Using feature vectors allows efficient feature extraction and improves classification performance [34].

It specifically uses pre-trained (Vgg16, ResNet50) trained to classify the 1000 image categories of the ImageNet dataset as the base for the proposed models. Using these pre-trained models' prior knowledge to get the basic features to be concatenated to the proposed model's second branch (MobileNet V2 or effecientnet) with its layers are trainable to extract the features of the natural and computer-colorized images dataset.

They demonstrated stacked ensemble learning by employing the two deep neural network models on the same dataset. The outputs of these models were then concatenated and fed as input to the classification phase level-1 Layer. This approach allows for exploiting the strengths of the two models, leading to improved performance compared to a single model's results. Using a pre-trained feature vector of either (VGG16 or ResNet50) combined with trainable MobileNet v2 or effecientnet feature vector help reduces the computational resources and time required for training these model from scratch, making the proposed models achieve the best accuracy and generalization performance while keeping the resources required to the minimum.

As Vgg16 achieved the best result, using Vgg16 with the efficientnet [36] feature vector setting its layers to trainable to check for the most accurate and generalization architecture for this classification problem.

### 3.3. Classification Phase

The classification phase uses one dense layer with two neurons, which acts as the level-1 layer that takes its inputs from the level-0 outputs. This output is the ensemble learning model consisting of the frozen pre-trained (Vgg16 or ResNet50) extracted features and the trainable MobilenetV2 or efficientnet's extracted features. These two results of extracted features enable the proposed model's level-1 layer to accurately learn to classify images using the two models' output to the labeled classes (natural and computer-colorized images).

The whole framework architecture for the proposed model 1 algorithm is shown below.

---
**Algorithm 1: Proposed Model 1 training (VGG16, mobilenet_v2)**

---
**Input:** Preprocessed datasets contain colorized and original color images
**Output:** Label the result for the detection (Colorized or Original)
1: Tr_DS,Val_DS = DS.split
2: Create a normalization rescaling layer (1.0/255)
3: Map the normalization layer to Tr_DS , Val_DS
4: Resize image to 224x224
5: Shuffle
6: Cache and prefetch Tr_DS & Val_DS
7: Vgg_model = load_vgg16_model_ with initial weights of imagenet
8: Set all vgg_model layers trainable property to false
9: Mob_model = load_mobilenet_v2_model _with initial weights of imagenet
10: Set all mob_model layers trainable property to True
11: inputs = create_input_layer
12: Vgg_output = flatten_vgg_output(vgg_model, inputs)
13: Mob_output = get_mobilenet_v2_output(mob_model, inputs)
14: x = concatenate_features(vgg_output, mob_output) concatenated along the last axis
15: outputs = Dense layer with two neurons with its input features (x)
16: Proposed_model_1 = create_composite_model(inputs, outputs)
17: Train the proposed model 1 using Preprocessed Tr_DS , Val_DS

---

For the proposed model 2, the VGG16 was replaced with Resnet50. For the proposed model 3, the mobilenet_v2 was replaced with efficientnet.

### 4. EXPERIMENT AND RESULTS

#### 4.1. Dataset:

Using Y.Guo et al. [13] dataset, the most commonly used dataset for evaluating the three proposed models, is the best approach to compare our models' performance with other models classifying nature and auto-colorized images. This dataset contains DS1, DS2, and DS3. DS1 and DS2, each of approximately 40000 images distributed as:

1) Ten thousand natural (10000) images were selected from the 50000 images ImageNet [24] validation dataset.
2) Thirty thousand (30000) images that have been auto-colorized from the grayscale version of the 10000 images in one- by the three state-of-the-art different colorization methods [9]-[11].

DS1 and DS2 datasets are each comprised of 40,000 images; consequently, the entire DS1 and DS2 datasets have almost 80,000 images.

DS3 comprises 5063 images of the Oxford buildings dataset [35] and their three-colorization image versions of the automatically colorized approaches [9]-[11], with a total of 20252 images for DS3.

### 4.2. Experimental Results

In this study, a rigorous experimental design was employed to ensure the validity of the results. The training and testing samples were carefully selected to avoid overlapping. This is crucial in avoiding overfitting and ensuring the model generalizes well to unseen data.

Using the Tensorflow pipelining technique to prefetch part of the DS1 dataset with 11997 images to train the three proposed models. This small part of the dataset is almost a third of the total DS1 size. Shuffling our dataset files, making this training sample distribution 3106 for the natural color images and 8898 for the three-colorization methods, with a natural to-colorized images ratio of 0.349. The Validation dataset natural to colorized images ratio is 0.243.

The proposed model 1 training accuracy achieves 99.74%, with a validation accuracy of 98.49%.

Testing the proposed model 1 using DS2 ~40000 images (another 10000 images of ImageNet with its three auto-colorized versions resulting from the three colorization methods), keeping the same class distribution of (3 colorized images -To - 1 raw image) the same distribution as DS1. The resultant accuracy was 96.52%.

Tested proposed model 1 using the DS3 dataset. The resultant accuracy was 97.16%.

Testing proposed model 1 accuracy using the remaining DS1 27991 images (40000 images excluding the previously ~12000 images used for model training) and then testing using DS2 dataset ~40000 images and then testing with DS3. All datasets have the same class distribution (three colorized images to one natural). The best resultant accuracy was of DS1 testing at 98.93%.

Using the proposed model 2, with ReseNet50 pre-trained feature vector branch instead of VGG16, keeping all the other hyper-parameters as it is, repeating the same training steps using the same datasets followed with the proposed model 1, the training accuracy was 99.68%, and the validation accuracy was 97.49%.

The proposed model 3 (with the Efficientnet model branch) used the same training and testing approaches.

Proposed models were evaluated using Accuracy (higher is better) and HTER (lower is better).

The HTER metric used can be calculated from the equation:

$$\text{HTER} = 0.5 * (\frac{FP}{TN+} + \frac{FN}{TP+FN})$$

TP, TN, FP, and FN are true positive, true negative, false positive, and false negative values, respectively.

The proposed models used a new training approach, using a set of 1 to 3 natural to colorized images that are colorized using three different colorization methods [9]-[11]. The proposed models, after training, can classify colorized images resulting from the state-of-the-art auto-colorization methods with high accuracy.

The proposed models are built on the pre-trained models with the initial weights of the ImageNet dataset and fine-tuned using a part of the DS1, which comprises images from the ImageNet dataset and three colorization versions for each image. The results of testing using another dataset images from Oxford buildings [25] dataset and their colorized versions (same colorization algorithms used with ImageNet) shows high classification performance.

The results of testing the three proposed models are listed in Table (4); the bold result indicates the best performance.

*Table 4 lists the three proposed models' results when tested using DS1, DS2, and DS3.*

| Dataset | DS1 | | DS2 | | DS3 | |
|---|---|---|---|---|---|---|
| Metrics | HTER | Accuracy | HTER | Accuracy | HTER | Accuracy |
| **Proposed Model 1** | 0.017 | 98.93% | **0.067** | **96.52%** | **0.058** | **97.16%** |
| **Proposed Model 2** | 0.033 | 98.23% | 0.146 | 92.64% | 0.124 | 93.75% |
| **Proposed Model 3** | **0.013** | **99.13%** | 0.103 | 94.72% | 0.107 | 94.55% |

From the results in Table 4, the proposed model 3 has the best performance when the training and testing using images from the DS1. In contrast, the proposed model 1 (Vgg16-mobilenet) has the best performance considering the generalization performance (when the training and testing datasets are different).

Proposed Model (1), (2), and (3) Confusion Matrices (C.M.) are shown below:

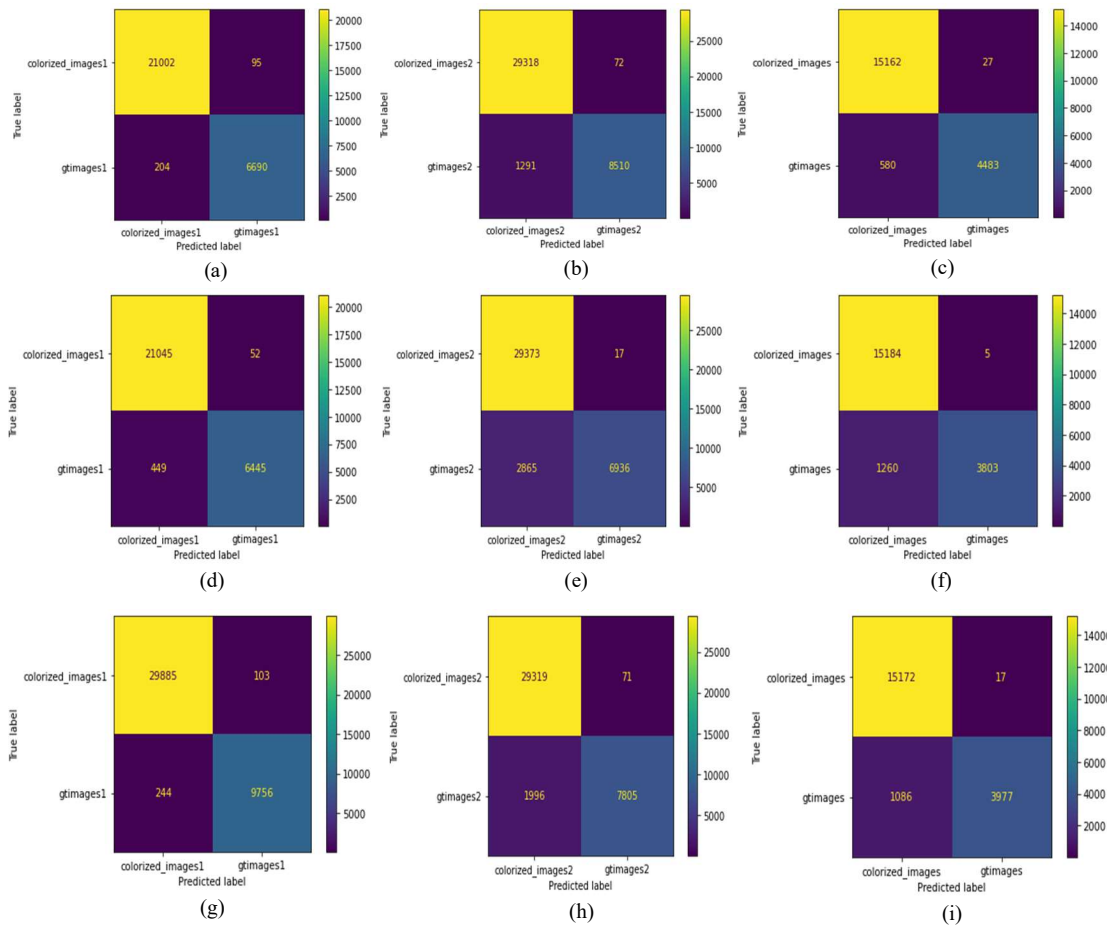

Figure 4. (a) Model 1 DS1 Test C.M, (b) Model 1 DS2 Test C.M., (c) Model 1 DS3 Test C.M.,
(d) Model 2 DS1 Test C.M, (e) Model 2 DS2 Test C.M, (f) Model 2 DS3 Test C.M
(g) Model 3 DS1 Test C.M, (h) Model 3 DS2 Test C.M, (i) Model 3 DS3 Test C.M

## 5. DISCUSSIONS:

From the results of the three proposed models, proposed model 1 had both the best classification accuracy and the best generalization performance. Accordingly, it will be used to be compared to the other models.

The results of the proposed model 1 from Table (4) and Table (3) are compared using internal validation HTER results testing the model using DS1 and external validation HTER results testing the model using DS2.

Shashikala s. et al. [42] compared their model with the state-of-the-art models used to detect auto-colorized images; in the following table (5), their results and this research's proposed model 3 will be compared in terms of accuracy and HTER. The proposed model 3 was selected from this research's proposed models as it achieves the best accuracy and the lowest HTER when tested using the same dataset used for its training which is the case with Shashikala s. et al. [42]'s proposed model gives an insight into the best overall model's accuracy.

*Table 5 shows the proposed model 3 accuracy and HTER results (in percentage)
Compared to Shashikala s. et al. [42] results.*

| Model | Accuracy | HTER |
|---|---|---|
| Shashikala s. et al. [42] | 0.949 | 18.21 |
| Li et al [18] | 0.927 | 20.84 |
| Guo et al [13] | 0.921 | 21.5 |
| Ulloa et [21] | 0.923 | 23.2 |
| Proposed Model 3 | 0.991 | 1.3 |

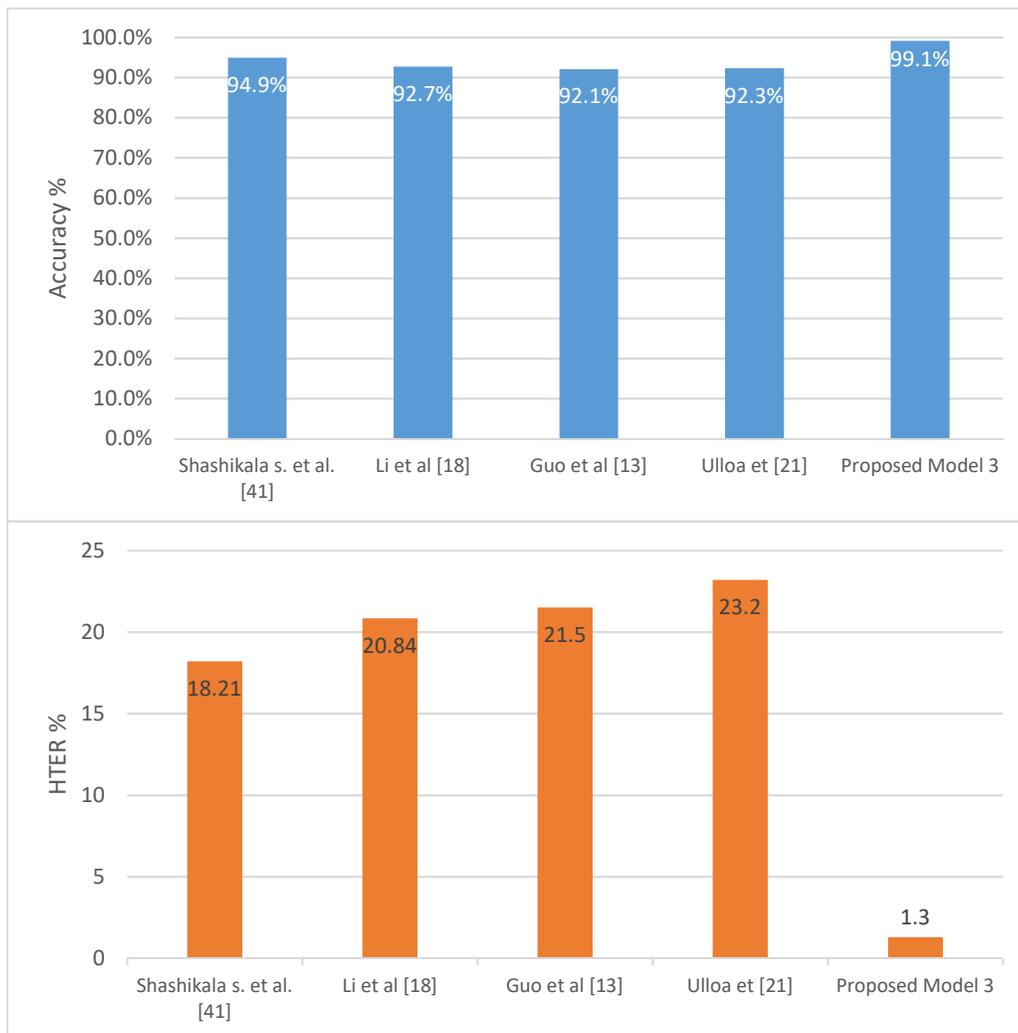

*Diagram (1) shows the proposed model 3 Accuracy and HTER (in percentage)
Compared with the Shashikala s. et al. [42] results.*

Diagram (1) shows the difference between the proposed model 3 HTER and Shashikala s. et al. [42] , Li et al. [18] , Guo et al. [13] , and Ulloa et al. [21] models, The comparison used only proposed model 3 as it is the best of the three proposed models in terms of classification accuracy and HTER.

Proposed model 1 achieves an HTER of 0.017 when tested using non-seen data from the same image dataset that was colorized using the same colorization method. It also reaches an HTER of 0.058 when tested using non-seen data from a different image dataset colorized using the same method. Then proposed model 1 outperforms all the related work models with an excellent generalization performance. Although proposed model 1 is trained using 1/3 of the DS1 dataset images, it achieves lower HTER, indicating higher accuracy.

It was considered that most of the related work papers used to evaluate their models using 2000 natural images with their corresponding 2000 colorized images. The proposed model was evaluated using datasets larger than ten times the subsets used for most related work methods, ensuring superior performance.

In the critical assessment of the proposed models, the potential impact of dataset variations affects the proposed models' performance and generalization capabilities. Some of the related work models used a cross-validation methodology where the dataset is divided into subsets for training and testing that introduces a potential difference in the distribution of subsets used in the cross-validation of this research's models. Considering the variations in dataset division and distribution used by those related work papers challenges the direct comparisons between their results and the proposed models. Nevertheless, our proposed models demonstrate exceptional accuracy and generalization capabilities when tested using the same dataset with a higher number of images and the same distribution as that used for its training.

## 6. CONCLUSION AND FUTURE WORK:

This study uses ensemble and transfer learning techniques to develop end-to-end models for detecting auto-colorized images. The proposed models achieve the best overall accuracy and generalization performance compared to previously published works in this area. The commonly used datasets for training and testing colorized image detection models have been used, which enable better comparisons with the proposed models.

Proposed Model 1, utilizing vgg16 and mobilenet_v2 branches, achieves a slightly lower detection accuracy than Proposed Model 3. However, it shows exceptional generalization performance with an accuracy of 97%. As Proposed Model 3 achieves the highest classification accuracy of 99.13% when using the same dataset for training and testing. The evaluation metrics employed in this study include accuracy and half-total error rate (HTER).

The findings of this research show the efficiency of incorporating ensemble learning and transfer learning approaches. The proposed models show high accuracy in detecting auto-colorized images that have been colorized using state-of-the-art colorization methods.

In future work, the plan is to test the proposed models by evaluating their performance in detecting example and scribble-based colorization methods and test them against the new fake colorized image datasets. This assessment will provide valuable insights into the generalization capabilities of our models and their potential to address various colorization techniques. Further investigations and discussions on the impact of dataset variations to better understand the limitations of our proposed models.